\documentclass{article}

\usepackage{arxiv}

\usepackage[utf8]{inputenc} 
\usepackage[T1]{fontenc}    
\usepackage{hyperref}       
\usepackage{url}            
\usepackage{booktabs}       
\usepackage{amsfonts}       
\usepackage{nicefrac}       
\usepackage{microtype}      
\usepackage{lipsum}
\usepackage{graphicx}

\usepackage{algorithm}
\usepackage{algpseudocode}

\title{Reinforcement Learning from Hierarchical Critics}

\author{
  Zehong Cao \\
  University of Tasmania, Australia\\
  \texttt{zhcaonctu@gmail.com} \\
   \And
  Chin-Teng Lin\\
  University of Technology Sydney, Australia\\
  \texttt{chin-teng.lin@uts.edu.au} \\
}

\begin{document}
\maketitle

\begin{abstract}
In this study, we investigate the use of global information to speed up the learning process and increase the cumulative rewards of reinforcement learning (RL) in competition tasks. Within the actor-critic RL, we introduce multiple cooperative critics from two levels of the hierarchy and propose a reinforcement learning from hierarchical critics (RLHC) algorithm. In our approach, each agent receives value information from local and global critics regarding a competition task and accesses multiple cooperative critics in a top-down hierarchy. Thus, each agent not only receives low-level details but also considers coordination from higher levels, thereby obtaining global information to improve the training performance. Then, we test the proposed RLHC algorithm against the benchmark algorithm, proximal policy optimisation (PPO), for two experimental scenarios performed in a Unity environment consisting of tennis and soccer agents' competitions. The results showed that RLHC outperforms the benchmark on both competition tasks.
\end{abstract}

\keywords{Reinforcement Learning; Hierarchy; Critics; Competition}

\section{Introduction}
Many agent training studies concern reinforcement learning (RL) techniques, which provide learning policies to achieve cooperative or competitive tasks by maximising rewards through interactions with the environment \cite{busoniu2010multi}. At each training step, the agent perceives the state of the environment and takes an action that causes the environment to transition into a new state. In a competitive game with multiple players, such as a zero-sum game for two agents, the mini-max principle, in which each player tries to maximise its benefits under the worst-case assumption that the opponent will always endeavour to minimise that benefit, is applied. For example, the minimax-Q algorithm \cite{littman2001value} employs the minimax principle to compute strategies and values for the stage games and a temporal-difference rule similar to Q-learning to propagate the values across state transitions. For more complex competition environments, such as StarCraft II tasks, \cite{rashid2018qmix} proposed a joint value-based method, QMIX, to coordinate between the centralised and decentralised policies. Furthermore, \cite{lowe2017multi} presented an adaptation of actor-critic methods that combines value-based methods in the critic and policy gradient methods in the actor. Following this, \cite{foerster2018counterfactual} recently proposed a new actor-critic method called counterfactual multi-agent (COMA) policy gradients that uses a centralised critic to estimate the Q value and decentralised actors to optimise the agents' policies. 

However, the above studies did not improve hierarchical learning with actor-critic structure. Considering the way in which agents are hierarchically structured may enable RL algorithms to overcome the challenges of excessive training time \cite{mnih2015human}. Inspired by feudal reinforcement learning \cite{dayan1993improving}, the DeepMind group proposed the feudal network (FuN) \cite{vezhnevets2017feudal}, which employs manager and worker modules for hierarchical reinforcement learning. The manager sets abstract goals, which are conveyed to and enacted by the worker, who generates primitive actions at each environment tick. The FuN structure has been extended to cooperative reinforcement learning \cite{ahilan2019feudal}, in which the manager learns to communicate sub-goals to multiple workers. Indeed, the ability to extract subgoals from the manager allows FuN to dramatically outperform a strong baseline agent on tasks.

Within the hierarchical structured RL, current RL methods as mentioned above focus more on assigning the subgoals and ignore the critical fact that giving the agent access to multiple cooperative critics might speed up the learning process and increase the rewards on competition tasks. In particular, it is frequently the case that high-level agents agree to be assigned different observations that work in combination with low-level agents to benefit hierarchical cooperation. Thus, in this study, we introduce multiple cooperative critics from two levels of the hierarchy and propose a reinforcement learning from hierarchical critics (RLHC) algorithm. The main contributions of our proposed approach are the following: (1) An agent receives information from both local and global critics regarding a competitive task. (2) The agent receives not only low-level details but also global information to consider coordination from higher levels to increase operational performance. (3) We define multiple cooperative critics in the top-to-bottom hierarchy, called reinforcement learning from hierarchical critics (RLHC). We assume that RLHC is a potential generalised RL and is thus more applicable for speeding up the training and improving the learning for agents. These benefits could potentially be obtained when using any type of hierarchical RL algorithm.

The remainder of this paper is organised as follows. In Section 2, we introduce the RL background for developing the multiple cooperative critic framework in agent competition domains. Section 3 describes the baseline and proposes the RLHC algorithm. Section 4 presents two experimental designs based on Unity-based tennis and soccer tasks with observation settings. Section 5 reports the training performance results of the benchmark algorithm and the proposed RLHC algorithm. Finally, we summarise the paper in Section 6.

\section{Preliminaries}
\subsection{Revisiting Reinforcement Learning}
In a standard RL framework \cite{kaelbling1996reinforcement}, an agent interacts with the external environment over a number of time steps. Here, $s$ is the set of all possible states, and $a$ is the set of all possible actions. At each time step $t$, the agent in state $s_t$ perceives the observation information $O_t$ from the environment, takes an action $a_t$, and receives feedback from the reward source $R_t$. Then, the agent transitions to a new state $s_{t+1}$, and the reward $R_{t+1}$ associated with the transition $(s_t, a_t, s_{t+1})$ is determined. The agent can choose an action from the last state visited. The goal of a reinforcement learning agent is to collect the maximum possible reward with minimal delay.

Next, we revisit the primary components in the learning process: MDP 
and the policy gradient.

In MDP, a state $S_t$ is Markov if and only if
\begin{equation}
\mathbb{P}[S_{t+1}|S_t] = \mathbb{P}[S_{t+1}|S_1,..., S_t]
\end{equation}
The future state is independent and unrelated to the past states. The state transition matrix $P$ is defined to present the transition probabilities from all states $s$ to all subsequent states $s'$.
\begin{equation}
P_{ss'}=\mathbb{P}[S_{t+1}=s'|S_t=s]
\end{equation}
A Markov reward process is a tuple $<S, A, P, R, \gamma>$, where $S$ is a finite set of states, $A$ is a finite set of actions, and $\gamma$ is a discount factor, $\gamma \in [0, 1]$.
\begin{itemize}
\item $P$ is a state transition probability matrix from Equation (2), \begin{equation}
P_{ss'}^{a} = \mathbb{P}[S_{t+1} = s'| S_t = s, A_t = a].
\end{equation}

\item $r$ is a reward function that represents the expected reward after the transition from $P$, \begin{equation}
r_s^a = \mathbb{E}[r_{t+1}|S_t=s, A_t = a]
\end{equation}\end{itemize}
The return $R_t$, defined as the sum of future discounted rewards, \begin{equation}
R_t=\sum_{k=0}^{\infty} \gamma^k r_{t+k+1}.
\end{equation}
To estimate ``how good'' it is to be in a given state, the state value function of the reward $V_\pi (s)$ is defined as the expected return starting with state $s$ under policy $\pi$
\begin{equation}
V_\pi (s) = \mathbb{E} [R_t|S_t = s, \pi],
\end{equation}
where policy $\pi$ $$\pi (a|s) = \mathbb{P}[A_t = a|S_t=s]$$.
Although the state value function suffices to define optimality, it is useful to define the action value of the reward function $Q_\pi (s, a)$:
\begin{equation}
Q_\pi (s, a) = \mathbb{E} [R_t|S_t = s, A_t = a, \pi].
\end{equation}Following the introduction of the value function, we can generate a gradient ascent-based RL, called the policy gradient. As a gradient ascent strategy, it models and optimises the policy directly. The policy is usually modelled by a parameterised function with respect to $\theta$,
$\pi_\theta (s, a)$. The value of the reward function depends on this policy and various other algorithms, such as REINFORCE (Monte Carlo Policy Gradient) \cite{silver2009monte}, deep deterministic policy gradient (DDPG) \cite{lillicrap2015continuous}, and asynchronous advantage actor-critic (A3C) \cite{mnih2016asynchronous}. Proximal policy optimisation (PPO) \cite{schulman2017proximal} can be applied to optimise $\theta$ to acquire the greatest reward.

The fundamental reward function is defined as follows:
\begin{equation}
\ J(\theta)=\mathbb{E}_{\pi \theta}[\pi_\theta(s, a)Q^{\pi_\theta}(s, a)],
\end{equation}

and then the gradient is computed:

\begin{equation}
\bigtriangledown_\theta J(\theta)=\mathbb{E}_{\pi \theta}[\bigtriangledown_\theta log \pi_\theta(s, a)Q^{\pi_\theta}(s, a)].
\end{equation}
\subsection{Actor-critic}
The actor-critic strategy aims to take advantage of the best characteristics from both the value-based and policy-based approaches while eliminating all their drawbacks and underlies recent modern RL methods from A3C to PPO. To understand the learning strategies, the value function can help with policy updates, such as by reducing gradient changes in the original strategy gradient, which is what actor-critic methods do. Specifically, actor-critic methods consist of two models that can optionally share parameters: (a) a critic updates the value function parameters $w$, which could be an action-value function $Q_w(s, a) $ or a state value function $V_w(s) $; (b) the actor updates the policy parameters $\theta$ for $\pi_\theta (s, a)$ in the direction suggested by the critic.

\subsubsection{Asynchronous Advantage Actor-Critic (A3C) }
The A3C structure \cite{mnih2016asynchronous} can master a variety of continuous motor control tasks and learn general game exploration strategies purely from observations. A3C maintains a policy ($\pi_\theta (s_t, a_t)$) and an estimate of the value function ($ V (s_t;\theta _w)$). Thread-specific parameters are synchronised with the global parameters: $\theta^\prime = \theta$ and $w^\prime = w$. This variant of actor criticism can operate in the forward view and uses the same mix of $n$-step returns to update both the policy and the value function. 

The update reward function can be written as follows:

\begin{equation}
\bigtriangledown_{\theta'} J(\theta')=\bigtriangledown _{\theta'} log \pi_{\theta'} {( s_t, a_t )} \hat{A}{(s_t, a_t ; \theta, \theta_w)},
\end{equation}
where $\hat{A}$ is an estimate of the advantage function given by
\begin{equation}
\hat{A}{(s_t, a_t ; \theta, \theta_w)} =\sum_{i=0}^{k-1}{\gamma^i r^{t+i} + \gamma^k V(s_{t+k}; \theta )-V(s_t;\theta _w)},
\end{equation}
and $k$ varies from state to state and has an upper bound of $t_{max} $.

The parameters $\theta$ (of the policy) and $\theta _w$ (of the value function) are shared even when they are shown to be separate for generality. For example, a convolutional neural network has one softmax output for the policy $\pi_\theta (s_t, a_t)$ and one linear output for the value function $ V (s_t;\theta _w)$, and all its non-output layers are shared.

\subsubsection{Proximal Policy Optimisation (PPO) }
PPO \cite{schulman2017proximal} is a new family of policy gradient methods for reinforcement learning that alternate between sampling data through interactions with the environment and optimising a surrogate objective function using stochastic gradient ascent. PPO imposes the constraint by forcing $r(\theta')$ to remain within a small interval of approximately 1, specifically, $[1-\varepsilon, 1+\varepsilon]$, where $\varepsilon$ is a hyperparameter. The function $clip(r(\theta'), 1-\varepsilon, 1+\varepsilon) $ clips the ratio within $[1-\varepsilon, 1+\varepsilon]$.

The objective function measures the total advantage over the state visitation distribution and actions,
\begin{equation}
\ J(\theta')=\mathbb{E}[r(\theta')\hat{A}^{\theta}(s, a)],
\end{equation}
where $r(\theta') = \pi_{\theta'}(s, a) / {\pi_\theta(s, a)}$ represents the probability ratio between the new and old policies.

To approximately maximise each iteration, the ``surrogate'' objective function is as follows:

\begin{equation}
\ J(\theta')=\mathbb{E}[min(r(\theta'))\hat{A}^{\theta}(s, a), clip(r(\theta), 1-\varepsilon, 1+\varepsilon)\hat{A}^{\theta}(s, a)].
\end{equation}

\section{Method}
To propagate the critics in the hierarchies, we propose RLHC by considering multiple cooperative critics in two levels of the hierarchy. RLHC aims to speed up the learning process and increase the cumulative rewards, as we assign each agent to receive information from both local and global critics. The novelty of this study is that it supports the concept that considering information from multiple critics at different levels is beneficial for training in a hierarchical reinforcement learning framework. The assumption is that a higher-level critic will be beneficial for an agent who was previously able to use only the critic in its surrounding layer. Thus, we address the modified advantage function performed by the maximum function in a union set based on the baseline, the benchmark PPO algorithm.

\subsection{Baseline: the Benchmark PPO}
PPO performs comparably to or better than other state-of-the-art RL methods and became the benchmark reinforcement learning algorithm at OpenAI \footnote{https://openai.com/blog/openai-baselines-ppo} and Unity \footnote{https://github.com/Unity-Technologies/ml-agents/blob/master/docs/Training-PPO.md} due to its ease of use and good performance. Here, we use PPO as both a baseline to validate the experiments and as a starting point to develop a novel RLHC algorithm.

\subsection{Learning in Multicritics}
To apply PPO to the problem of agents with variable attention to more than one critic, we consider the argument for resolving the multiple-critic learning problem. For each critic $i$, the corresponding advantage function is $A^{\theta_i}(s_i, a)$ generated from the state value function $V^{i}(s_i, \theta)$, depending on the different scale observations $O$ (expressed as the state $s$) and the network parameter $\theta$. Consistent with existing work, the advantage function is extended from the value function and measures the value of the agent's actions.

For multiple critics (such as two critics representing $i =2$), we work with the argument of the minimum objective function to find the minimum advantage that represents the benefit of choosing a specific action instead of following the current policy. The argument of the minimum objective function can be written as follows:
\begin{equation}
argmin(\bigtriangledown_J(\theta'))=arg min(\mathbb{E}[\bigtriangledown_r(\theta')\hat{A}^{\theta}(s, a)]).
\end{equation}

To achieve a minimised $\hat{A}^{\theta}(s, a)$, we need to maximise the current state value function $V(s;\theta )$ extracted from Equation (11), which can be written as

\begin{equation}
min[\hat{A}(s, a)] \rightarrow max[V(s;\theta)],
\end{equation}

in other words, the set of $\hat{V_i}^{\theta}$ of the given argument of objective function $J(\theta')$ for which the value of the given expression attains its maximum value. Because the maximum $\hat{V}{(s, \theta)} $ indicates that action $a$ is a better choice than the current policy $\pi(\theta)$, we measure the advantage function performed by collecting individual $\hat{V}^{i}(s, \theta)$ and choosing the maximum $\hat{V}{(s, \theta)} $. The corresponding updated value function can be written as follows:
\begin{equation}
\hat{V}{(s, \theta)} = max\bigcup_{i=2}^{m} \hat{V}^{i}(s, \theta),
\end{equation}where $m$ is the total number of critics.

If we consider the $n$ time step intervals of multiple critics, then $m$ in Equation (16) can be replaced with $h_t$, where $h_t=h_{t+kT}$, $h_t=m$ , $k=2, 3, 4,...$, and $T$ is a time period with $n$ time steps; otherwise, $h_t=2$.

\subsection{RLHC}
In terms of propagating the critics in the hierarchies, we are the first to develop an RL strategy from hierarchical critics allowing a worker agent $i$ to receive information from multiple critics computed both locally and globally. The manager is responsible for collecting the broader observations and estimating the corresponding global critic, which it sends to the worker agent. To clarify our proposed algorithm, we also show the pseudo-code of our proposed RLHC below.

\begin{algorithm}
\caption{RLHC}
\begin{algorithmic}[1]

\State{Define the observation environment for each worker agent $i$ $\to O_w^i$ and manager $\to O_m$}
\State{Initialise $\rightarrow$ the state of each worker agent $s_0^{w_i}$, the state of the manager $s_0^m$ and the policy parameter $\theta_0$ }
\State{Initialise $\rightarrow$ the critic networks of manager $\hat{A_0^m}$ and each worker $\hat{A_0^{w_i}}$ and the actor network for each worker $\pi_0^{w_i}$ to determine action $a$}

\For{Iteration $ = 1, 2,...$}
\For{Actor $ = 1, 2, ..., i$}

\State{Run policy $\pi_\theta$ in the environment for $T$ $\in t$ time steps}
\State{Using $\theta'$ to interact with the environment to collect ${s_t, a_t}$ and compute advantage function $\hat{A_t^\theta}$}
\State{Minimise the gradient of the objective function $\bigtriangledown_ J(\theta')$}
\State{$\to$ measure the probability ratio $r(\theta') $ between new and old policies}
\State{$\to$ find the maximum current value function $\hat{V_t^\theta}$ to achieve the minimum advantage function $\hat{A_t^\theta}$ selected from the advantage estimate of worker agent $i$ or the manager}
$$argmin\bigtriangledown_ J(\theta') =$$ $$\mathbb{E}[min[(\bigtriangledown_r^w(\theta')\hat{A_t}^{\theta}(s_t^{w_i}, a_t), \bigtriangledown_r^m(\theta')\hat{A_t}^{\theta}(s_t^m, a_t)] $$

$$\rightarrow max[V(s_t^{w_i};\theta), V(s_t^{m};\theta)]$$

\State{Choose the maximum value:}
$$\hat{V_t^\theta} = max\bigcup_{i=1} (V(s_t^{w_i};\theta), V(s_t^{m};\theta)) $$
\State{Use the maximum value $\hat{V_t^\theta}$ to calculate the advantage estimates $\hat{A_1^\theta}$,..., $\hat{A_T^\theta}$ }

\EndFor
\State{Optimise the ``surrogate'' objective function from PPO }
\State{Update $\theta' \to \theta$}
\EndFor
\label{code:recentStart}
\end{algorithmic}
\end{algorithm}

Here, we apply the RLHC algorithm in the PPO. The successfully trained RLHC model requires tuning of the trained hyperparameters, which is beneficial for the output of the training process containing the optimised policy. This investigation allows criticism from the manager to improve the training performance.

Furthermore, we draw Fig. 1 as a simplification to demonstrate the RLHC algorithm constructed using a two-level hierarchy for one worker agent with a manager. The local and global critics are implemented by the maximum function illustrated in the ``learning in multicritics''. For the modified state value function we propose, the manager and worker share the actors, but they provide different critics from the two layers (we consider it hierarchical), which correspond to the arrows and the maximum function in Fig. 1. The manager receives the shared action space from the worker but provides only high-level criticism to the worker. This strategy not only allows us to estimate the value of multiple critics from different levels but also further allows the use of weighted approaches to fuse critics from different layers or to optimise the temporal scaling of critics in separate layers. For simplicity, the experiments in the following section presented in this study generally use two-level hierarchies, such as a multi-agent hierarchy with up to 2 managers and 4 worker agents for competition.

\begin{figure}[H]
\centering\includegraphics [width=8cm, height=7.5cm] {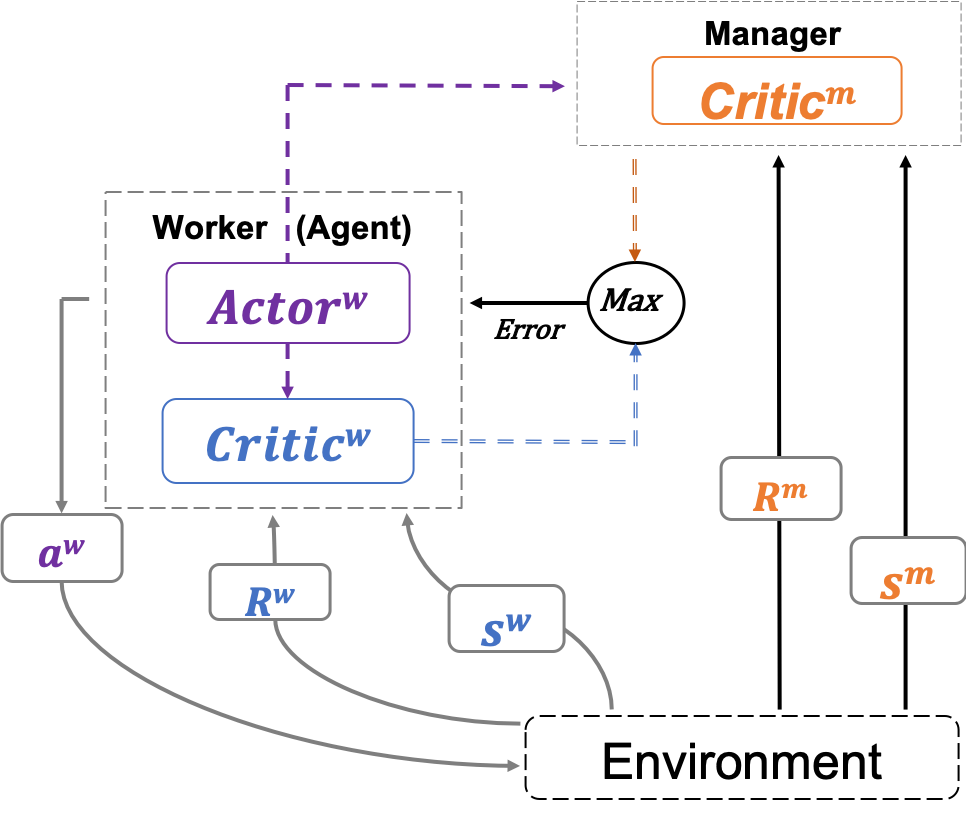} 
\centerline{Figure 1. The RLHC algorithm}
\end{figure}

\section{Experiment}
We applied our proposed RLHC algorithm to two scenarios in which up to 4 agents compete. We empirically show the success of our RLHC compared with the benchmark PPO method in competitive scenarios such as tennis and toy 
soccer. We have released codes for both the model and the environments on GitHub for replication purposes.

\subsection{Unity Platform for RL}
Because many existing platforms (e.g., OpenAI Gym) lack the ability to configure the simulation for multiple agents flexibly, the simulation environment becomes a black box from the perspective of the learning system. The Unity platform, a new open-source toolkit, has been developed for creating and interacting with simulation environments. Specifically, the Unity machine learning agents toolkit (ML-Agents Toolkit) \cite{juliani2018unity} is an open-source Unity plugin that enables games and simulations to serve as environments for training multiple intelligent agents. The toolkit supports dynamic multi-agent interaction, and agents can be trained using RL through a straightforward Python API.

\subsection{Scenario 1: Tennis Competition}
In this game, agents control rackets to bounce the ball over a net. We constructed a new training environment in Unity under 2-2 workers and 1-1 manager settings (a doubles-tennis scenario) as shown in Fig. 2. Referring to Table 1, the goal, agent reward function, and behaviour parameters, including action and observation spaces, are set up for the tennis agents. Please note we set extended local individual observations, where the low-level agents (racket workers) can also access the distance and velocity difference of teammates to avoid duplicated policies and actions. The manager observations include additional variables, such as the distance between the ball and the racket and information gained from the worker agent's observations. 

\begin{table}\footnotesize
\centering
\begin{tabular}{p{0.1\textwidth}p{0.35\textwidth}}
\textbf{Setting} & \textbf{Description} \\
\hline
Objective & Agents shall not miss the ball or let the ball fall out of the court area during the episode by striking the ball over the net into the opponents' court. \\
\hline
Reward & +0.1 when the ball is hit over the net \\
 & -0.1 when agents miss the ball or the ball falls out of the tennis court \\
\hline
Action Space &  The movement forward or away from the net as well as jumping (3 variables). \\
\hline
Observation Space & Position and velocity information of the ball and racket (8 variables). \\
\hline
Manager Observation & Position and velocity information of the ball and racket and the distance between the ball and racket (10 variables)\\
\hline
Observation Space - Extended & Position and velocity information of the ball and racket and the distance and velocity difference of teammates (12 variables) \\
\hline
Manager Observation - Extended & Position and velocity information of the ball and racket, the distance between the ball and racket, and the distance and velocity difference of teammates (14 variables)\\
\hline
\end{tabular}
\caption{Settings of the tennis competition scenario}
\end{table}

\begin{figure}[H]
\centering\includegraphics [width=8.5cm, height=5cm] {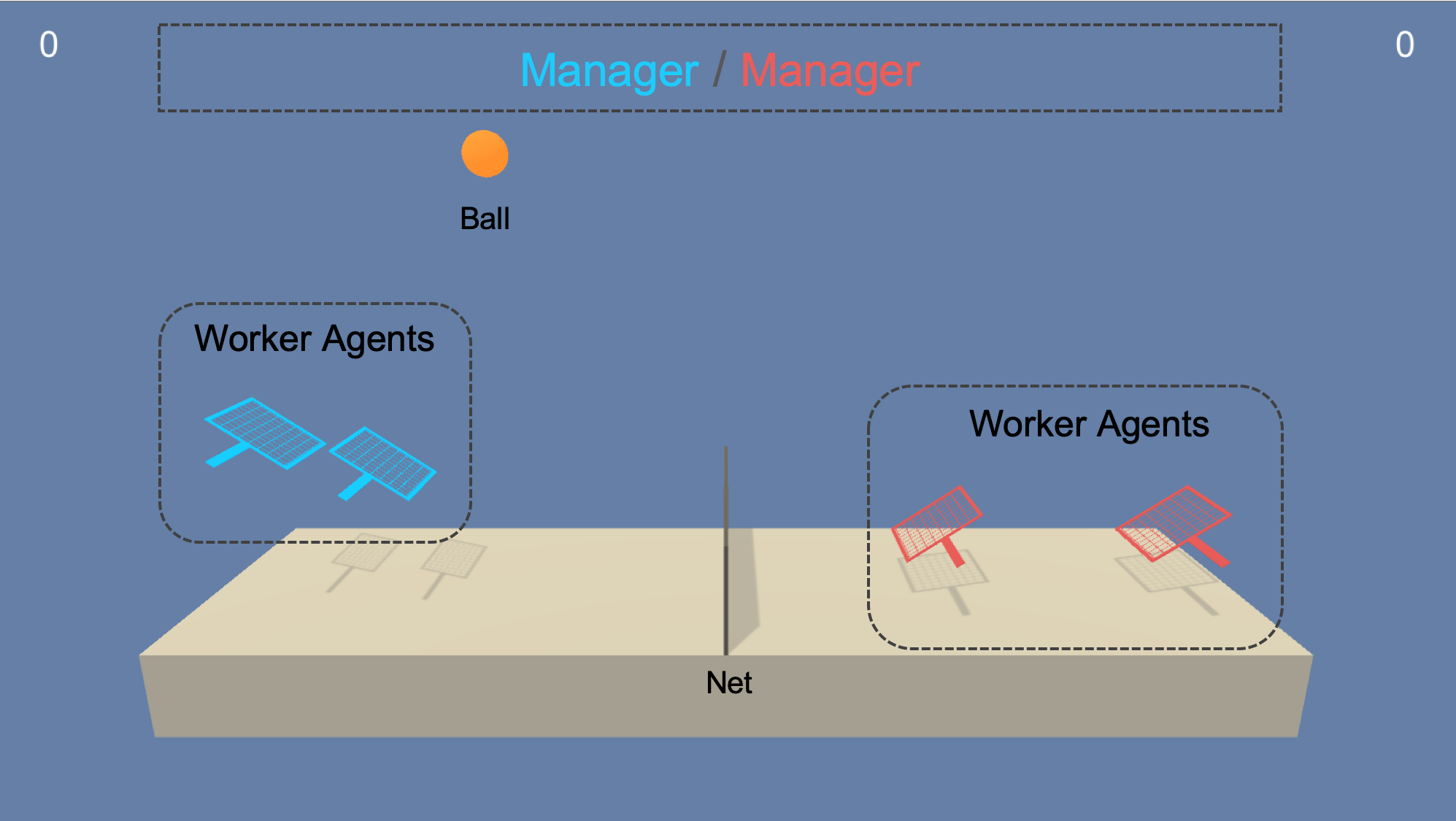} 
\centerline {Figure 2 Tennis competition 2 vs. 2 in Unity}
\end{figure}

\subsection{Scenario 2: Soccer Competition}
This competition is an environment where 4 agents compete in a simplified soccer game in Unity. Fig. 3 shows the environment where 4 agents compete in a 2 vs. 2 soccer game. This game has two types of players, offence and defence, which need to be controlled differently. We use ``multibrain training'' in Unity because each team contains one striker agent and one goalie agent, and each is trained using separate reward functions; thus, each type has its own observation and action space. As presented in Table 2, the goals, agent reward function, and behaviour parameters, including action and observation spaces, are set up for the soccer agents.

\begin{table}\footnotesize
\centering
    \begin{tabular}{p{0.1\textwidth}p{0.35\textwidth}}
        \textbf{Setting} & \textbf{Description} \\
        \hline
        Objective & Striker agents need to calculate a method to kick the ball into the opponent's goal. \\
        & Goalie agents need to learn to defend against the opponent and to avoid the ball being kicked into their own goal. \\
        \hline
        Reward & Striker: +1 when the ball enters the opponent's goal, -0.1 when the ball enters own team's goal. \\
        & Goalie: -1 when the ball enters own team's goal, +0.1 when the ball enters the opponent's goal. \\
        \hline
        Action Space &  Striker: Forward, backward, rotation and sideways movement (6 variables) \\
        & Goalie: Forward, backward and sideways movement (4 variables)\\
        \hline
        Observation Space & Seven types of object detection, with distance information in 180 degrees of view (112 variables)\\
        \hline
        Manager Observation & Eight types of object detection, with distance information in 270 degrees of view (200 variables) \\
        \hline
    \end{tabular}
    \caption{Settings of the soccer competition scenario}
\end{table}

\begin{figure}[H]
\centering\includegraphics [width=8.5cm, height=5cm] {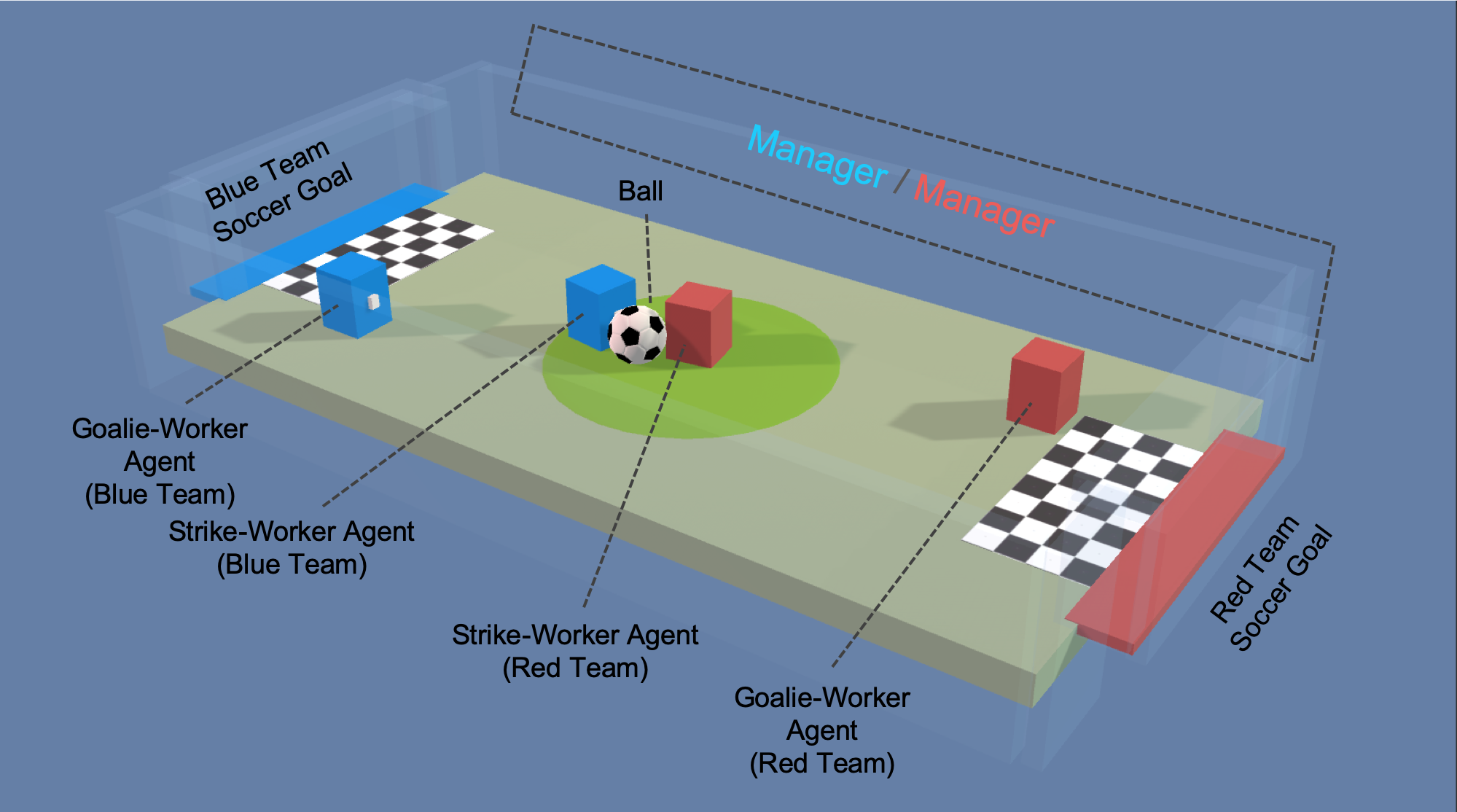} 
\centerline {Figure 3 Soccer competition 2 vs. 2 in Unity}
\end{figure}

\subsection{Training Settings and Metrics}
\subsubsection{Training Settings}
The hyper-parameters for the RL used for training are specified in Table 3, which provides the initialisation settings that we used to interact with the tennis or soccer competition environment. Specifically, the batch size and buffer size represent the number of experiences that occur during each gradient descent iteration and the number of experiences to collect before updating the policy model, respectively. Beta controls the strength of entropy regularisation, and epsilon influences how rapidly the policy can evolve during training. Gamma and lambda indicate the reward discount rate for the generalised advantage estimator and the regularisation parameter, respectively.

\begin{table}\footnotesize
\centering 
\label{tab:plain}
\begin{tabular}{lll|lll}
\hline

\hline
& Tennis & Soccer & & Tennis & Soccer \\
\hline
Parameters & Values & Values & Parameters & Values & Values\\

\hline
batch size & 1024 & 128 & beta & 0.005 & 0.01 \\
buffer size & 10240 & 2000 & epsilon & 0.2 & 0.2 \\
gamma & 0.99 & 0.99 & hidden units & 128 & 256 \\
lambda & 0.95 & 0.95 & learning rate & 0.0003 & 0.001 \\
max steps & 200 K & 500 K & memory size & 256 & 256 \\
normalise & true & false & num. epoch & 3 & 3 \\
num. layers & 2 & 2 & time horizon & 64 & 128 \\
sequence len. & 64 & 64 & summary freq. & 1000 & 2000 \\
\hline
\end{tabular}
\caption{Settings of training parameters}
\end{table}

\subsubsection{Training Metrics}
We saved some statistics during the learning session and viewed them using a TensorFlow utility named TensorBoard. Here, we measure four metrics to assess training performance. Specifically, \textit{cumulative reward} indicates the mean cumulative episode reward accrued by all agents interacting with the environment. \textit{Episode length} is the mean length of each episode in the environment for all agents in that environment. \textit{Entropy} controls the degree of randomness of model decisions. \textit{Value} estimates the mean value estimate for all states visited by the agent.

\section{Results}
We provide the training performances of the RLHC algorithm and the baseline benchmark algorithm (PPO). PPO uses an independent local critic for each agent and does not share information, thus rendering the environment nonstationary from a single-agent's perspective. However, our RLHC includes a semicentralised critic by hierarchically assigning a critic to estimate the updated value function; this can be beneficial for independent learners, which are known to struggle in hierarchically cooperative settings.

The following findings show that RLHC is both more efficient and more general than PPO; consequently, we choose two example scenarios for use with 4-player tennis and soccer competitions. To study the training process in more detail, we use TensorBoard (with smoothing $= 0.8$) to demonstrate the cumulative reward, episode length, entropy, and value estimate for the training metrics.

\subsection {Tennis Competition}
For the tennis competition (the doubles scenario), we use both standard and extended observations for training purposes. As shown in Table 1, we set 2 observation space categories, worker and manager, consisting of the (standard) observations and extended observations, respectively, to determine whether the extended observation is beneficial in achieving a higher reward. We also compare the performance metrics' of RLHC and the benchmark PPO in terms of both the standard and extended observations.

As shown in Fig. 4, considering the standard observations, RLHC achieves a higher cumulative reward and a longer episode length with short training steps compared with PPO. Furthermore, after adding the extended observations, the cumulative reward trained by PPO further increases compared with RLHC without extended observations, indicating that the extended observations that consider teammate relationships are significant in the training process. Additionally, we include the extended observations in RLHC and PPO to compare their training performances. The metrics, cumulative reward and episode length show that our RLHC achieves better performances than PPO. Similarly, the value estimate increases rapidly in our RLHC compared with PPO. Both methods provide a successful training process, and both present slowly decreasing entropy.

\begin{figure}[H]
\centering\includegraphics [width=9cm, height=7cm] {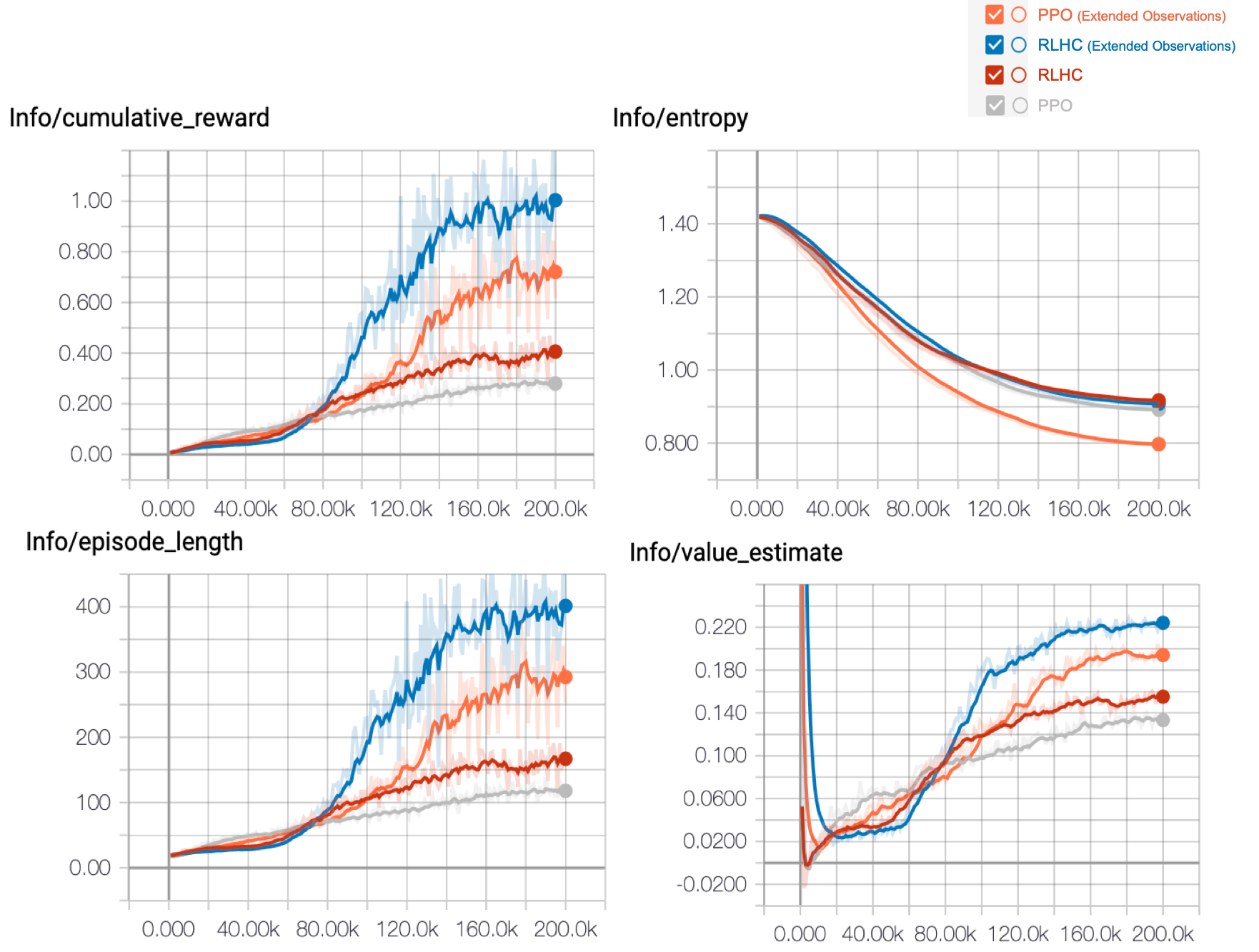} 
\centerline{Figure 4. The training metrics for tennis competition} 
\end{figure}

\subsection {Soccer Competition}
For the soccer competition, we set the observation spaces for the worker and the manager to assess a different view, as shown in Table 2. During the training stage, we trained both brains: one brain with a negative reward for the ball entering their goal and another brain with a positive reward for the ball entering the opponent's goal. As the mean reward will be inverse between the striker and goalie and crisscrosses 
during training, we only demonstrate the training metrics for the striker agent, as shown in Fig. 5. The corresponding training metrics for the goalie agent are inversed from the striker agent.

In terms of the striker's performance, Fig. 5 shows that the cumulative reward in PPO increased around the starting points and then decreased after 200K training steps, suggesting this trial does not have 
a reliable learning process. However, our RLHC can achieve a positive result with higher cumulative rewards compared with PPO. Moreover, the episode length in PPO keeps rising due to a possibly unstable learning process, but the episode length in RLHC is stable after 60K training steps. Additionally, the value estimate of our RLHC increases and converges after 60K training steps.

\begin{figure}[H]
\centering\includegraphics [width=9.5cm, height=7.5cm] {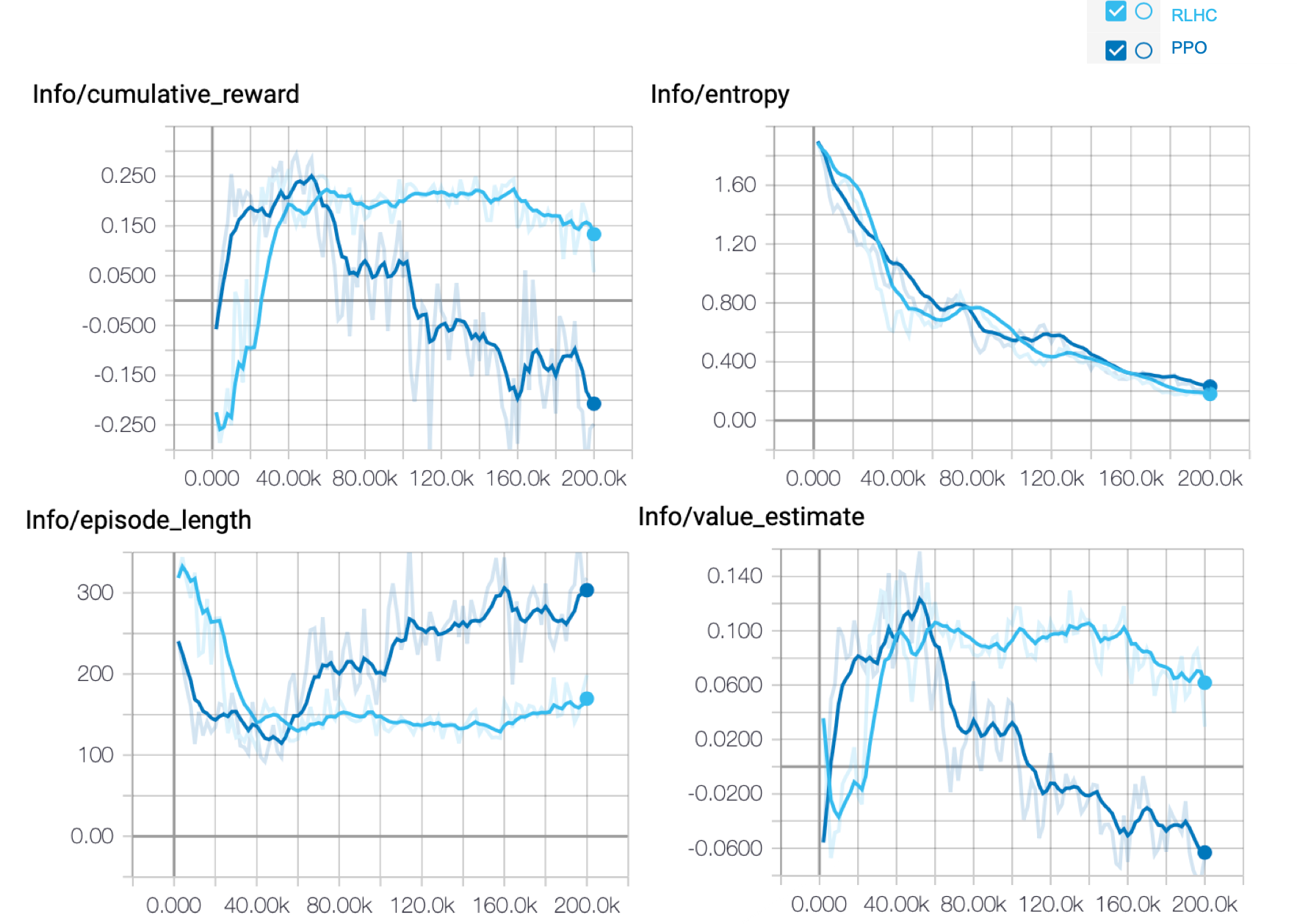} 
\centerline{Figure 5. The striker's training metrics for the soccer competition}
\end{figure}

\section{Conclusion}
In this study, we developed the RLHC algorithm to consider global information to speed up the learning process and increase the cumulative rewards. Within RLHC, the agent is allowed to receive information from both local and global critics in competitive tasks. We tested the proposed RLHC on two tasks, 4-player tennis and soccer competition, in the Unity environment by comparing its results with those of the benchmark PPO algorithm. The results showed that our proposed RLHC outperforms the nonhierarchical critic baseline PPO on agent-competition tasks. The novelty of this study is that it shows a proof-of-concept that considering multiple critics from different levels can be beneficial for training in a hierarchical RL framework. We selected a simple scenario as evidence, and the preliminary outcomes showed improved performance by considering the criticism from the higher-level critics.

\bibliographystyle{unsrt}  
\bibliography{references}


\end{document}